\documentclass[hidelinks]{article} 
\usepackage{iclr2018_conference,times}
 
\usepackage[utf8]{inputenc} 
\usepackage[T1]{fontenc}    
\usepackage{hyperref}       
\usepackage{url}            
\usepackage{booktabs}       
\usepackage{amsfonts}       
\usepackage{nicefrac}       
\usepackage{microtype}      
\usepackage{xcolor}
\usepackage{epsfig}
\usepackage{isomath}
\usepackage{graphicx}
\usepackage{tabularx}
\usepackage{amsmath}
\usepackage{amssymb}
\usepackage{multirow}
\usepackage{bbm}
\usepackage{color}
\usepackage{mathtools}
\usepackage{bm}
\usepackage{textcomp}
\usepackage{footnote}
\usepackage{subcaption}
\usepackage{booktabs}
\usepackage{bigstrut}
\usepackage{wrapfig}
\usepackage{enumitem}
\iclrfinalcopy

\usepackage[linesnumbered,algoruled,boxed,lined]{algorithm2e}

\newenvironment{algotabularx}
 {\tabularx{\linewidth-\inoutsize-\widthof{~~~}-4\tabcolsep-\rightskip}[t]}
 {\endtabularx}

\DeclarePairedDelimiter\ceil{\lceil}{\rceil}
\DeclarePairedDelimiter\floor{\lfloor}{\rfloor}

\graphicspath{{./figures/}}

\title{Accelerating Neural Architecture Search using Performance Prediction}

\newif\ifnote
\notetrue

\DeclarePairedDelimiterX{\infdivx}[2]{(}{)}{%
  #1\;\delimsize\|\;#2%
}

\author{
  Bowen Baker$^1$$^*$, Otkrist Gupta$^1$$^*$, Ramesh Raskar$^1$,   Nikhil Naik$^2$ \\
  (* - indicates equal contribution) \\
  $^1$\textbf{MIT Media Laboratory}, Cambridge, MA 02139\\
  $^2$\textbf{Harvard University}, Cambridge, MA 02138\\
  $\{$bowen, otkrist, raskar, naik$\}$@mit.edu \\
}

\begin{document}

\maketitle

\begin{abstract}
Methods for neural network hyperparameter optimization and meta-modeling are computationally expensive due to the need to train a large number of model configurations. In this paper, we show that standard frequentist regression models can predict the final performance of partially trained model configurations using features based on network architectures, hyperparameters, and time-series validation performance data. We empirically show that our performance prediction models are much more effective than prominent Bayesian counterparts, are simpler to implement, and are faster to train. Our models can predict final performance in both visual classification and language modeling domains, are effective for predicting performance of drastically varying model architectures, and can even generalize between model classes. Using these prediction models, we also propose an early stopping method for hyperparameter optimization and meta-modeling, which obtains a speedup of a factor up to 6x in both hyperparameter optimization and meta-modeling. Finally, we empirically show that our early stopping method can be seamlessly incorporated into both reinforcement learning-based architecture selection algorithms and bandit based search methods. Through extensive experimentation, we empirically show our performance prediction models and early stopping algorithm are state-of-the-art in terms of prediction accuracy and speedup achieved while still identifying the optimal model configurations.
\end{abstract}
\section{Introduction}
\vspace{-5pt}
At present, significant human expertise and labor is required for designing high-performing neural network architectures and successfully training them for different applications. Ongoing research in two areas---meta-modeling and hyperparameter optimization---attempts to reduce the amount of human intervention required for these tasks. Hyperparameter optimization methods (e.g.,~\cite{hutter2011sequential,snoek2015scalable,li2016hyperband}) focus primarily on obtaining good optimization hyperparameter configurations for training human-designed networks, whereas meta-modeling algorithms~\citep{bergstra2013making,verbancsics2013generative,baker2016designing,zoph2016neural} aim to design neural network architectures from scratch. Both sets of algorithms require training a large number of neural network configurations for identifying the right set of hyperparameters or the right network architecture---and are hence computationally expensive. 

When sampling many different model configurations, it is likely that many subpar configurations will be explored. Human experts are quite adept at recognizing and terminating suboptimal model configurations by inspecting their partial learning curves. In this paper we seek to emulate this behavior and automatically identify and terminate subpar model configurations in order to speedup both meta-modeling and hyperparameter optimization methods. Our method parameterizes learning curve trajectories with simple features derived from model architectures, training hyperparameters, and early time-series measurements from the learning curve. We use these features to train a set of frequentist regression models that predicts the final validation accuracy of partially trained neural network configurations using a small training set of fully trained curves from both image classification and language modeling domains. We use these predictions and uncertainty estimates obtained from small model ensembles to construct a simple early stopping algorithm that can speedup both meta-modeling and hyperparameter optimization methods.

While there is some prior work on neural network performance prediction using Bayesian methods~\citep{domhan2015speeding,klein2017learning}, our proposed method is significantly more accurate, accessible, and efficient. We hope that our work leads to inclusion of neural network performance prediction and early stopping in the practical neural network training pipeline. 
\section{Related Work}
\vspace{-5pt}
\label{sec:related}
\textbf{Neural Network Performance Prediction:}
There has been limited work on predicting neural network performance during the training process. \cite{domhan2015speeding} introduce a weighted probabilistic model for learning curves and utilize this model for speeding up hyperparameter search in small convolutional neural networks (CNNs) and fully-connected networks (FCNs). Building on~\cite{domhan2015speeding},~\cite{klein2017learning} train Bayesian neural networks for predicting unobserved learning curves using a training set of fully and partially observed learning curves. Both methods rely on expensive Markov chain Monte Carlo (MCMC) sampling procedures and handcrafted learning curve basis functions.
We also note that ~\cite{swersky2014freeze} develop a Gaussian Process kernel for predicting individual learning curves, which they use to automatically stop and restart configurations.

\textbf{Meta-modeling:} We define meta-modeling as an algorithmic approach for designing neural network architectures from scratch. The earliest meta-modeling approaches were based on genetic algorithms~\citep{schaffer1992combinations,stanley2002evolving,verbancsics2013generative} or Bayesian optimization~\citep{bergstra2013making,shahriari2016taking}. More recently, reinforcement learning methods have become popular.~\cite{baker2016designing} use Q-learning to design competitive CNNs for image classification.~\cite{zoph2016neural} use policy gradients to design state-of-the-art CNNs and Recurrent cell architectures.  Several methods for architecture search~\citep{pmlr-v70-cortes17a,negrinho2017deeparchitect,zoph2017learning,brock2017smash,suganuma2017genetic} have been proposed this year since the publication of~\cite{baker2016designing} and~\cite{zoph2016neural}.

\textbf{Hyperparameter Optimization:} We define hyperparameter optimization as an algorithmic approach for finding optimal values of design-independent hyperparameters such as learning rate and batch size, along with a limited search through the network design space. Bayesian  hyperparameter optimization methods include those based on sequential model-based optimization (SMAC)~\citep{hutter2011sequential}, Gaussian processes (GP)~\citep{snoek2012practical}, TPE~\citep{bergstra2013making}, and neural networks~\cite{snoek2015scalable}. 
However, random search or grid search is most commonly used in practical settings~\citep{bergstra2012random}. Recently,~\cite{li2016hyperband} introduced Hyperband, a multi-armed bandit-based efficient random search technique that outperforms state-of-the-art Bayesian optimization methods.

\section{Neural Network Performance Prediction}
\vspace{-5pt}
\label{sec:method}
We first describe our model for neural network performance prediction, followed by a description of the datasets used to evaluate our model, and finally present experimental results. \vspace{-5pt}

\begin{figure}
	\centering
	\includegraphics[width=\textwidth]{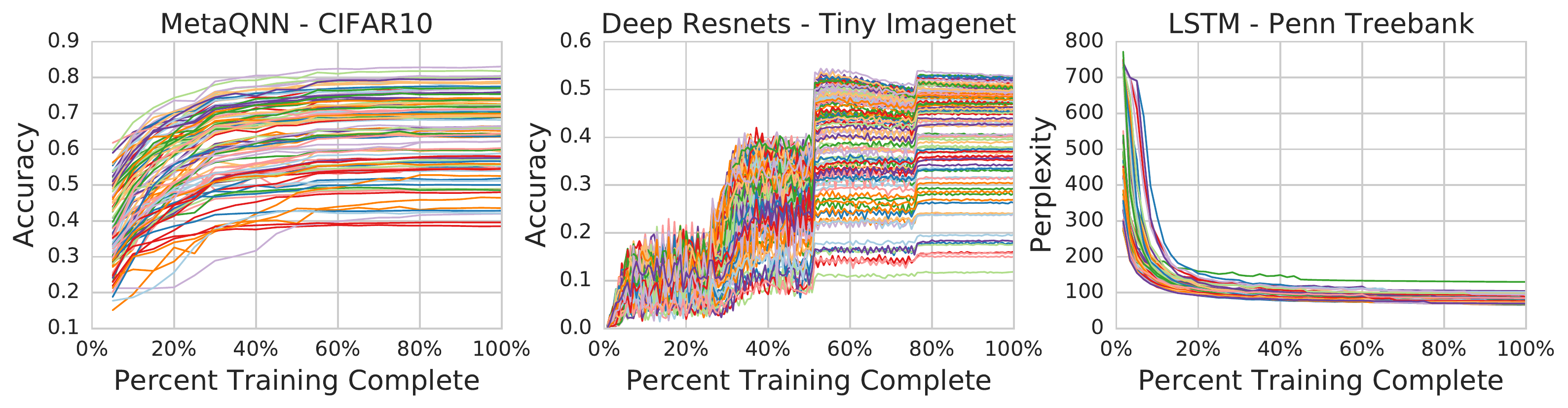}
	\caption{\textbf{Example Learning Curves:} Example learning curves from experiments considered in this paper. Note the diversity in convergence times and overall learning curve shapes.}
	\label{learning_curves}
\end{figure}

\subsection{Modeling Learning Curves}
\vspace{-5pt}
Our goal is to model the validation accuracy $y_T$ of a neural network configuration $\mathbf{x} \in \mathcal{X} \subset \mathbb{R}^d$ at epoch $T \in \mathbb{Z}^+ $ using previous performance observations $y(t)$. For each configuration $\mathbf{x}$ trained for $T$ epochs, we record a time-series $y(T) = y_1, y_2, \dots, y_T$ of validation accuracies. We train a population of $n$ configurations, obtaining a set $\mathcal{S}  = \{(\mathbf{x}^1, y^{1}(t)), (\mathbf{x}^2, y^{2}(t)), \ldots , (\mathbf{x}^n, y^{n}(t))\}$. Note that this problem formulation is very similar to ~\cite{klein2017learning}.

We propose to use a set of features $u_\mathbf{x}$, derived from the neural network configuration $\mathbf{x}$, along with a subset of time-series accuracies $y(\tau) = (y_t)_{t = 1, 2, \ldots, \tau}$ (where $1\leq \tau < T$) from $\mathcal{S}$ to train a regression model for estimating $y_T$. Our model predicts $y_T$ of a neural network configuration using a feature set $x_f = \{u_\mathbf{x}, y(t)_{\text{1--$\tau$}}\}$. For clarity, we train $T-1$ regression models, where each successive model uses one more point of the time-series validation data. As we shall see in subsequent sections, this use of \textit{sequential regression models} (SRM) is more computationally and more precise than methods that train a single Bayesian model. 

\textbf{Features:} We use features based on time-series (TS) validation accuracies, architecture parameters (AP), and hyperparameters (HP). (1) TS: These include the validation accuracies $y(t)_{\text{1--$\tau$}} = (y_t)_{t = 1, 2, \ldots, \tau}$ (where $1\leq \tau < T$), the first-order differences of validation accuracies (i.e., ${y_t}' = (y_t - y_{t-1})$), and the second-order differences of validation accuracies (i.e., ${y_t}'' = ({y_t}' - {y_{t-1}')}$). (2) AP: These include total number of weights and number of layers. (3) HP: These include all hyperparameters used for training the neural networks, e.g., initial learning rate and learning rate decay (full list in Appendix Table 2).

\subsection{Datasets and Training Procedures}
\vspace{-7pt}
We experiment with small and very deep CNNs (e.g., ResNet, Cuda-Convnet) trained on image classification datasets and with LSTMs trained with Penn Treebank (PTB), a language modeling dataset. Figure \ref{learning_curves} shows example learning curves from three of the datasets considered in our experiments. We provide brief summary of the datasets below. Please see Appendix Section \ref{appendix:search_space} for further details on the search space, preprocessing,  hyperparameters and training settings of all datasets. 

\textbf{Datasets with Varying Architectures:}

\textbf{Deep Resnets (TinyImageNet):} We sample 500 ResNet architectures and train them on the TinyImageNet\footnote{https://tiny-imagenet.herokuapp.com/} dataset (containing 200 classes with 500 training images of $32\times32$ pixels) for 140 epochs. We vary depths, filter sizes and number of convolutional filter block outputs. The network depths vary between 14 and 110. 

\textbf{Deep Resnets (CIFAR-10):} We sample 500 39-layer ResNet architectures from a search space similar to~\cite{zoph2016neural}, varying kernel width, kernel height, and number of kernels. We train these models for 50 epochs on CIFAR-10.  

\textbf{MetaQNN CNNs (CIFAR-10 and SVHN):} We sample 1,000 model architectures from the search space detailed by~\cite{baker2016designing}, which allows for varying the numbers and orderings of convolution, pooling, and fully connected layers. The models are between 1 and 12 layers for the SVHN experiment and between 1 and 18 layers for the CIFAR-10 experiment. Each architecture is trained on SVHN and CIFAR-10 datasets for 20 epochs.

\textbf{LSTM (PTB):} We sample 300 LSTM models and train them on the Penn Treebank dataset for 60 epochs, evaluating perplexity on the validation set. We vary number of LSTM cells and hidden layer inputs between 10-1400.  

\textbf{Datasets with Varying Hyperparameters:}

\textbf{Cuda-Convnet (CIFAR-10 and SVHN):} We train Cuda-Convnet architecture~\citep{cudaconv} with varying values of initial learning rate, learning rate reduction step size, weight decay for convolutional and fully connected layers, and scale and power of local response normalization layers. We train models with CIFAR-10 for 60 epochs and with SVHN for 12 epochs.

\subsection{Prediction Performance}
\vspace{-0.5pt}
\textbf{Choice of Regression Method:} We now describe our results for performing final neural network performance. 
For all experiments, we train our SRMs on 100 randomly sampled neural network configurations. We obtain the best performing method using random hyperparameter search over 3-fold cross-validation. We then compute the regression performance over the remainder of the dataset using the coefficient of determination $R^2$. We repeat each experiment 10 times and report the results with standard errors. We experiment with a few different frequentist regression models, including ordinary least squares (OLS), random forests, and $\nu$-support vector machine regressions ($\nu$-SVR). As seen in Table \ref{tab:freq}, $\nu$-SVR with linear or RBF kernels perform the best on most datasets, though not by a large margin. For the rest of this paper, we use $\nu$-SVR RBF unless otherwise specified. 
\begin{table}[]
\centering
\begin{tabular}{|l|l|l|l|l|}
\hline
Dataset         			& $\nu$-SVR (RBF)       	& $\nu$-SVR (Linear) 	& Random Forest 	& OLS\\ \hline
MetaQNN (CIFAR-10) 			& $94.22 \pm0.25$ 			& $94.44\pm0.14$ 		& $92.27\pm0.91$ 	& $93.22\pm1.1$			\\ \hline
Resnet (TinyImageNet) 		& $85.78 \pm1.82$ 			& $91.8\pm1.1$ 			& $91.37 \pm2.18$	& $90.15\pm1.8$ 		\\ \hline
LSTM (Penn Treebank) 		& $83.29\pm7.71$ 			& $98.59\pm0.8$ 		& $91.38\pm1.97$ 	& $89.8\pm0.16$			\\ \hline
\end{tabular}
\caption{\textbf{Frequentist Model Comparison:} We report the coefficient of determination $R^2$ for four standard methods. Each model is trained with 100 samples on 25\% of the learning curve. We find that $\nu$-SVR works best on average, though not by a large margin.}
\label{tab:freq}
\end{table}

\begin{figure}
\centering
\includegraphics[width=0.9\textwidth]{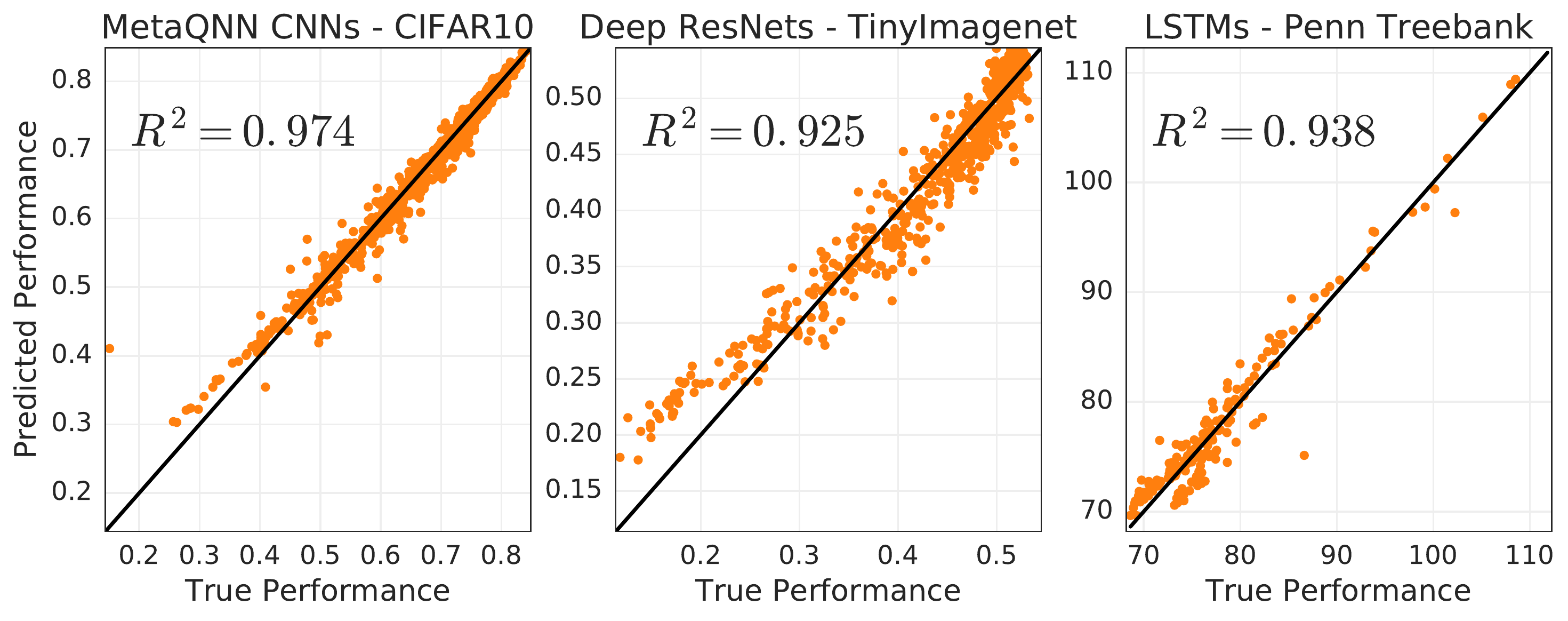}
\caption{\textbf{Predicted vs True Values of Final Performance:} We show the shape of the predictive distribution on three experiments: MetaQNN models, Deep Resnets, and LSTMs. Each $\nu$-SVR (RBF) model is trained with 100 configurations with data from 25\% of the learning curve. We predict validation set classification accuracy for MetaQNN and Deep ResNets, and perplexity for LSTMs.}
\label{scatter}
\end{figure}

\textbf{Ablation Study on Feature Sets:} In Table~\ref{tab:abl}, we compare the predictive ability of different feature sets, training SVR (RBF) with time-series (TS) features obtained from 25\% of the learning curve, along with features of architecture parameters (AP), and hyperparameters (HP). TS features explain the largest fraction of the variance in all cases. For datasets with varying architectures, AP are more important that HP; and for hyperparameter search datasets, HP are more important than AP, which is expected. AP features almost match TS on the ResNet (TinyImageNet) dataset, indicating that choice of architecture has a large influence on accuracy for ResNets. Figure~\ref{scatter} shows the true vs. predicted performance for all test points in three datasets, trained with TS, AP, and HP features. 

\textbf{Generalization Between Depths:} We also test to see whether SRMs can accurately predict the performance of out-of-distribution neural networks. In particular, we train SVR (RBF) with 25\% of TS, along with AP and HP features on ResNets (TinyImagenet) dataset, using 100 models with number of layers less than a threshold $d$ and test on models with number of layers greater than $d$, averaging over 10 runs. Value of $d$ varies from 14 to 110. For $d=32$, $R^2$ is $80.66 \pm 3.8$. For $d=62$, $R^2$ is $84.58 \pm 2.7$. 

\begin{table}[]
\centering
\begin{tabular}{|l|l|l|l|l|}
\hline
Feature Set & MetaQNN  & ResNets & LSTM & Cuda-Convnet\\
& (CIFAR-10) & (TinyImageNet) & (Penn Treebank) & (CIFAR-10)\\ \hline
TS 					& $93.98 \pm 0.15$ 			& $86.52 \pm 1.85$ 				& $97.81 \pm 2.45$ & $95.54 \pm 0.24$  		\\ \hline
AP 					& $27.45 \pm 4.25$ 			& $84.33 \pm 1.7$ 					& $16.11 \pm 1.13$ 	& $1.1 \pm 0.6$	\\ \hline
HP 					& $12.60 \pm 1.79$ 			& $8.78 \pm 1.14$ 					& $3.98 \pm 0.88$ & $18.19 \pm 2.19$			\\ \hline
TS+AP 				& $84.09 \pm 1.4$ 			& $88.82 \pm 2.95$ 				& $96.92  \pm  2.8$) & $95.36 \pm 0.27$		\\ \hline
AP+HP 				& $27.01 \pm  5.2$  		& $81.71 \pm 3.9$ 					& $15.97 \pm 2.57$ 	& $21.65 \pm 2.72$	\\ \hline
TS+AP+HP 			& $94.44 \pm 0.14$ 			& $91.8 \pm 1.1$ 					& $98.24  \pm 2.11$ & $95.69 \pm 0.15$		\\ \hline
\end{tabular}
\caption{\textbf{Ablation Study on Feature Sets:} Time-series features (TS) refers to the partially observed learning curves, architecture parameters (AP) refer to the number of layers and number of weights in a deep model, and hyperparameters (HP) refer to the optimization parameters such as learning rate. All results with SVR (RBF). 25\% of learning curve used for TS.}
\label{tab:abl}
\end{table}

\subsubsection{Comparison with Existing Methods:} 
\begin{figure*}[t]
\centering
\includegraphics[width=1\textwidth]{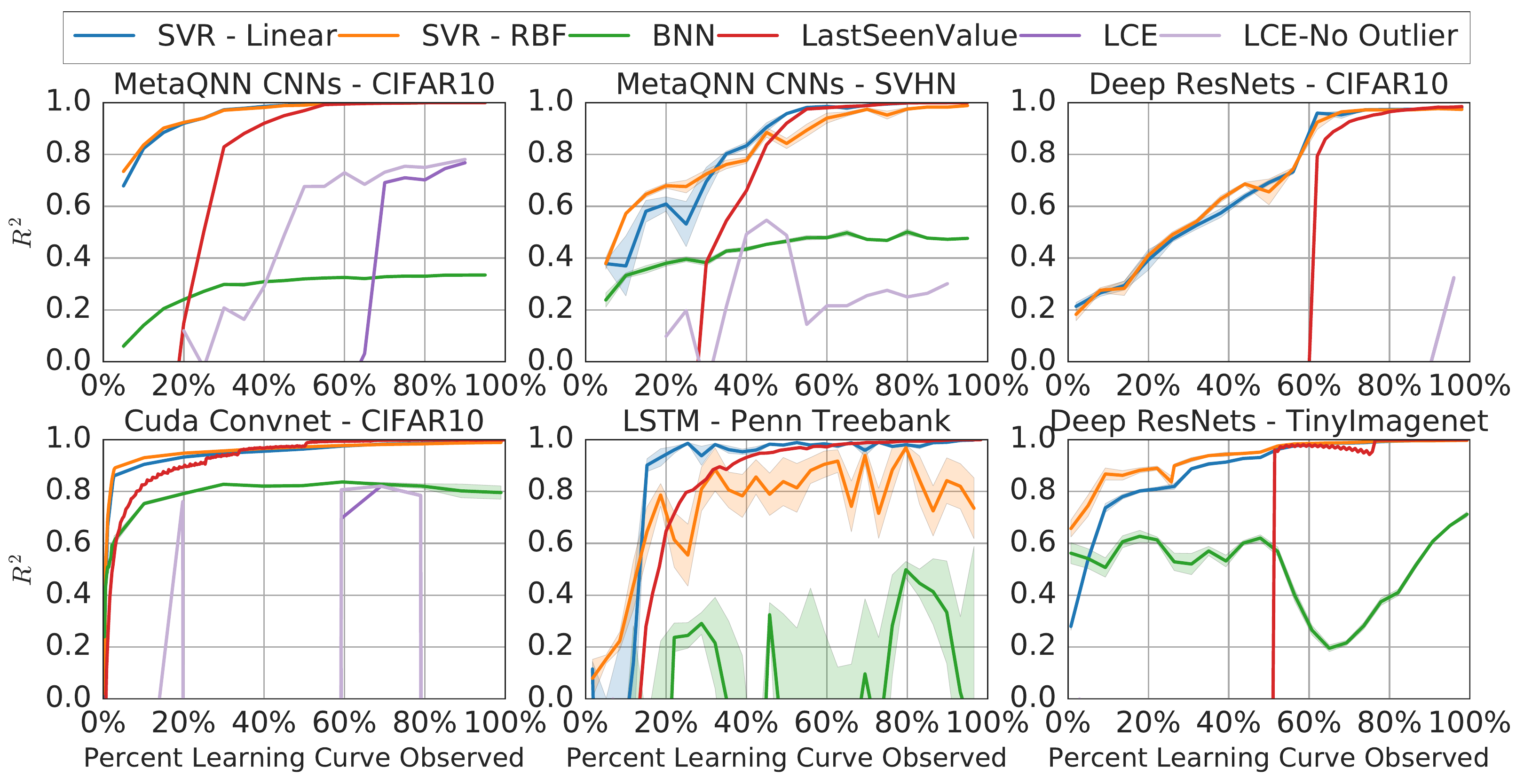}
\caption{\textbf{Performance Prediction Results:} We plot the performance of each method versus the percent of learning curve observed. For BNN and $\nu$-SVR (linear and RBF), we sample 10 different training sets, plot the mean $R^2$, and shade the corresponding standard error. We compare our method against BNN~\citep{klein2017learning}, LCE~\citep{domhan2015speeding}, and a ``last seen value'' heuristic \citep{li2016hyperband}. Absent results for a model indicate that it did not achieve a positive $R^2$. The results for Cuda-Convnet on the SVHN dataset are shown in Appendix Figure \ref{perf_svhn_cuda}.}
\label{results_by_percent}
\end{figure*}

\vspace{-0.7pt}
We now compare the neural network performance prediction ability of SRMs with three existing learning curve prediction methods: (1) Bayesian Neural Network (BNN)~\citep{klein2017learning}, (2) the learning curve extrapolation (LCE) method~\citep{domhan2015speeding}, and (3) the last seen value (LastSeenValue) heuristic~\citep{li2016hyperband}. When training the BNN, we not only present it with the subset of fully observed learning curves but also all other partially observed learning curves from the training set. While we do not present the partially observed curves to the $\nu$-SVR SRM for training, we felt this was a fair comparison as $\nu$-SVR uses the entire partially observed learning curve during inference.  Methods (2) and (3) do not incorporate prior learning curves during training.
Figure \ref{results_by_percent} shows the $R^2$ obtained by each method for predicting the final performance versus the percent of the learning curve used for training the model. We see that in all neural network configuration spaces and across all datasets, either one or both SRMs outperform the competing methods. The LastSeenValue heuristic only becomes viable when the configurations are near convergence, and its performance is worse than an SRM for very deep models. We also find that the SRMs outperform the LCE method in all experiments, even after we remove a few extreme prediction outliers produced by LCE. Finally, while BNN outperforms the LastSeenValue and LCE methods when only a few iterations have been observed, it does worse than our proposed method. In summary, we show that our simple, frequentist SRMs outperforms existing Bayesian approaches on predicting neural network performance on modern, very deep models in computer vision and language modeling tasks. 

Since most of our experiments perform stepwise learning rate decay; it is conceivable that the performance gap between SRMs and both LCE and BNN results from a lack of sharp jump in their basis functions. We experimented with exponential learning rate decay (ELRD), which the basis functions in LCE are designed for. We trained 630 random nets with ELRD, from the 1000 MetaQNN-CIFAR10 nets. Predicting from 25\% of the learning curve, the $R^2$ is 0.95 for $\nu$-SVR (RBF), 0.48 for LCE (with extreme outlier removal, negative without), and 0.31 for BNN. This comparison illuminates another benefit of our method: we do not require handcrafted basis functions to model new learning curve types.

\textbf{Training and Inference Speed Comparison:}
Another advantage of our regression approach is speed. SRMs are much faster to train and do inference in than proposed Bayesian methods~\citep{domhan2015speeding,klein2017learning}. On 1 core of a Intel 6700k CPU, an $\nu$-SVR (RBF) with 100 training points trains in 0.006 seconds, and each inference takes 0.00006 seconds. In comparison, the LCE code  takes 60 seconds and BNN code takes 0.024 seconds on the same hardware for each inference. 

\section{Applying Performance Prediction For Early Stopping}
\vspace{-5pt}
\label{sec:early_stop}
To speed up hyperparameter optimization and meta-modeling methods, we develop an algorithm to determine whether to continue training a partially trained model configuration using our sequential regression models. If we would like to sample $N$ total neural network configurations, we begin by sampling and training $n \ll N$ configurations to create a training set $\mathcal{S}$. We then train a model $f(x_f)$to predict $y_T$. Now, given the current best performance observed $y_\text{BEST}$, we would like to terminate training a new configuration $\mathbf{x'}$ given its partial learning curve $y'(t)_{\text{1--$\tau$}}$ if $f({x_f}') = \hat y_T \le y_\text{BEST}$ so as to not waste computational resources exploring a suboptimal configuration. 

However, in the case $f(x_f)$ has poor out-of-sample generalization, we may mistakenly terminate the optimal configuration. If we assume that our estimate can be modeled as a Gaussian perturbation of the true value $\hat y_T \sim \mathcal{N}(y_T, \sigma(\mathbf{x}, \tau))$, then we can find the probability $p(\hat y_T \le y_\text{BEST} | \sigma(\mathbf{x}, \tau)) = \Phi(y_\text{BEST}; y_T, \sigma)$, where $\Phi(\cdot; \mu, \sigma)$ is the CDF of $\mathcal{N}(\mu, \sigma)$. Note that in general the uncertainty will depend on both the configuration and $\tau$, the number of points observed  from the learning curve. Because frequentist models do not admit a natural estimate of uncertainty, we assume that $\sigma$ is independent of $\mathbf{x}$ yet still dependent on $\tau$ and estimate it via Leave One Out Cross Validation. 

Now that we can estimate the model uncertainty, given a new configuration $\mathbf{x'}$ and an observed learning curve $y'(t)_{\text{1--$\tau$}}$,  we may set our termination criteria to be $p(\hat y_T \le y_\text{BEST}) \ge \Delta$. $\Delta$ balances the trade-off between increased speedups and risk of prematurely terminating good configurations. In many cases, one may want several configurations that are close to optimal, for the purpose of ensembling. We offer two modifications in this case. First, one may relax the termination criterion to $p(\hat y_T \le y_\text{BEST} - \delta) \ge \Delta$, which will allow configurations within $\delta$ of optimal performance to complete training. One can alternatively set the criterion based on the $n^{\text{th}}$ best configuration observed, guaranteeing that with high probability the top $n$ configurations will be fully trained. 



\subsection{Early Stopping for Meta-modeling}
\vspace{-5pt}
\cite{baker2016designing} train a $Q$-learning agent to design convolutional neural networks. In this method, the agent samples architectures from a large, finite space by traversing a path from input layer to termination layer. 
However, the MetaQNN method uses 100 GPU-days to train 2700 neural architectures and the similar experiment by~\cite{zoph2016neural} utilized 10,000 GPU-days to train 12,800 models on CIFAR-10. The amount of computing resources required for these approaches makes them prohibitively expensive for large datasets (e.g., Imagenet) and larger search spaces. The main computational expense of reinforcement learning-based meta-modeling methods is training the neural network configuration to $T$ epochs (where $T$ is typically a large number at which the network stabilizes to peak accuracy).

\begin{figure}[t]
\centering
\includegraphics[width=0.9\textwidth]{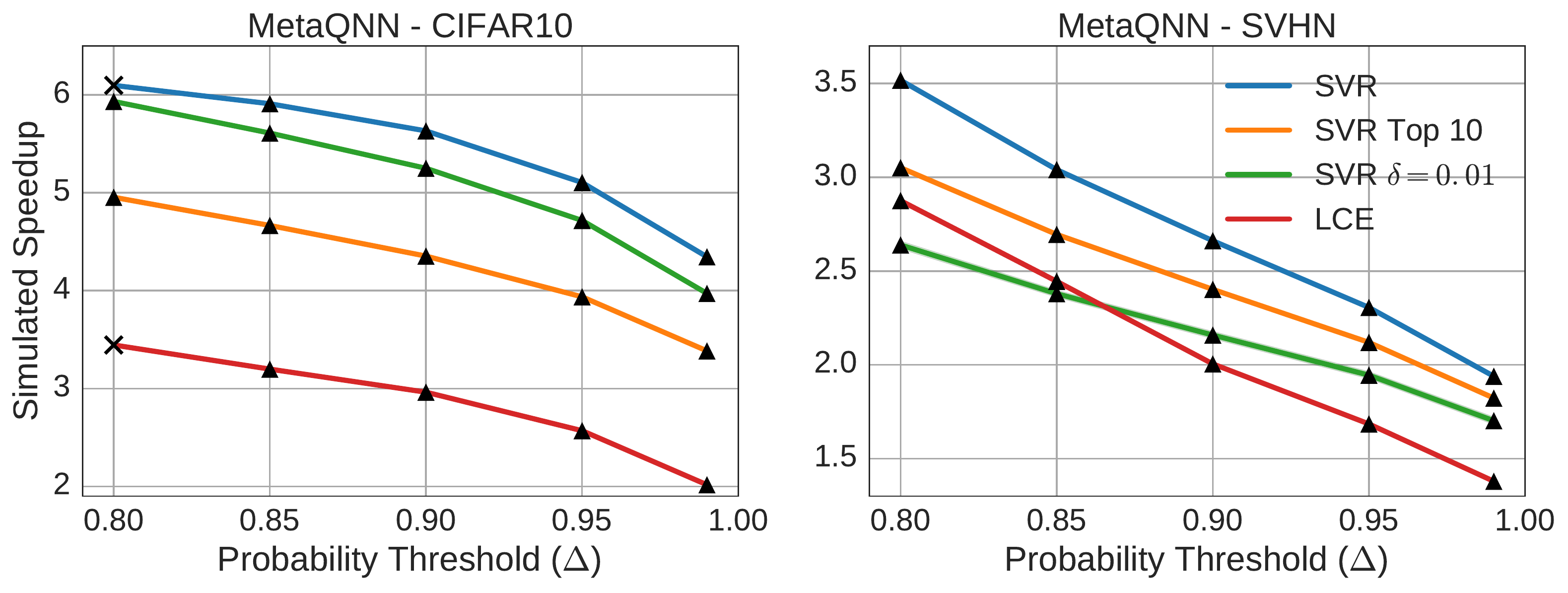}
\caption{\textbf{Simulated Speedup in MetaQNN Search Space:} We compare the three variants of the early stopping algorithm presented in Section \ref{sec:early_stop}. Each $\nu$-SVR SRM is trained using the first 100 learning curves, and each algorithm is tested on 10 independent orderings of the model configurations. Triangles indicate an algorithm that successfully recovered the optimal model for more than half of the 10 orderings, and X's indicate those that did not.}
\label{qnn_speed_up}
\vspace{-10pt}
\end{figure}

We now detail the performance of a $\nu$-SVR (RBF) SRM in speeding up architecture search using sequential configuration selection. First, we take 1,000 random models from the MetaQNN~\citep{baker2016designing} search space. We simulate the MetaQNN algorithm by taking 10 random orderings of each set and running our early stopping algorithm. We compare against the LCE early stopping algorithm~\citep{domhan2015speeding} as a baseline, which has a similar probability threshold termination criterion. Our SRM trains off of the first 100 fully observed curves, while the LCE model trains from each individual partial curve and can begin early termination immediately. Despite this ``burn in'' time needed by an SRM, it is still able to significantly outperform the LCE model (Figure \ref{qnn_speed_up}). In addition, fitting the LCE model to a learning curve takes between 1-3 minutes on a modern CPU due to expensive MCMC sampling, and  it is necessary to fit a new LCE model each time a new point on the learning curve is observed. Therefore, on a full meta-modeling experiment involving thousands of neural network configurations, our method could be faster by several orders of magnitude as compared to LCE based on current implementations. 

\begin{wrapfigure}{r}{0.45\textwidth}
\centering
\includegraphics[width=0.45\textwidth]{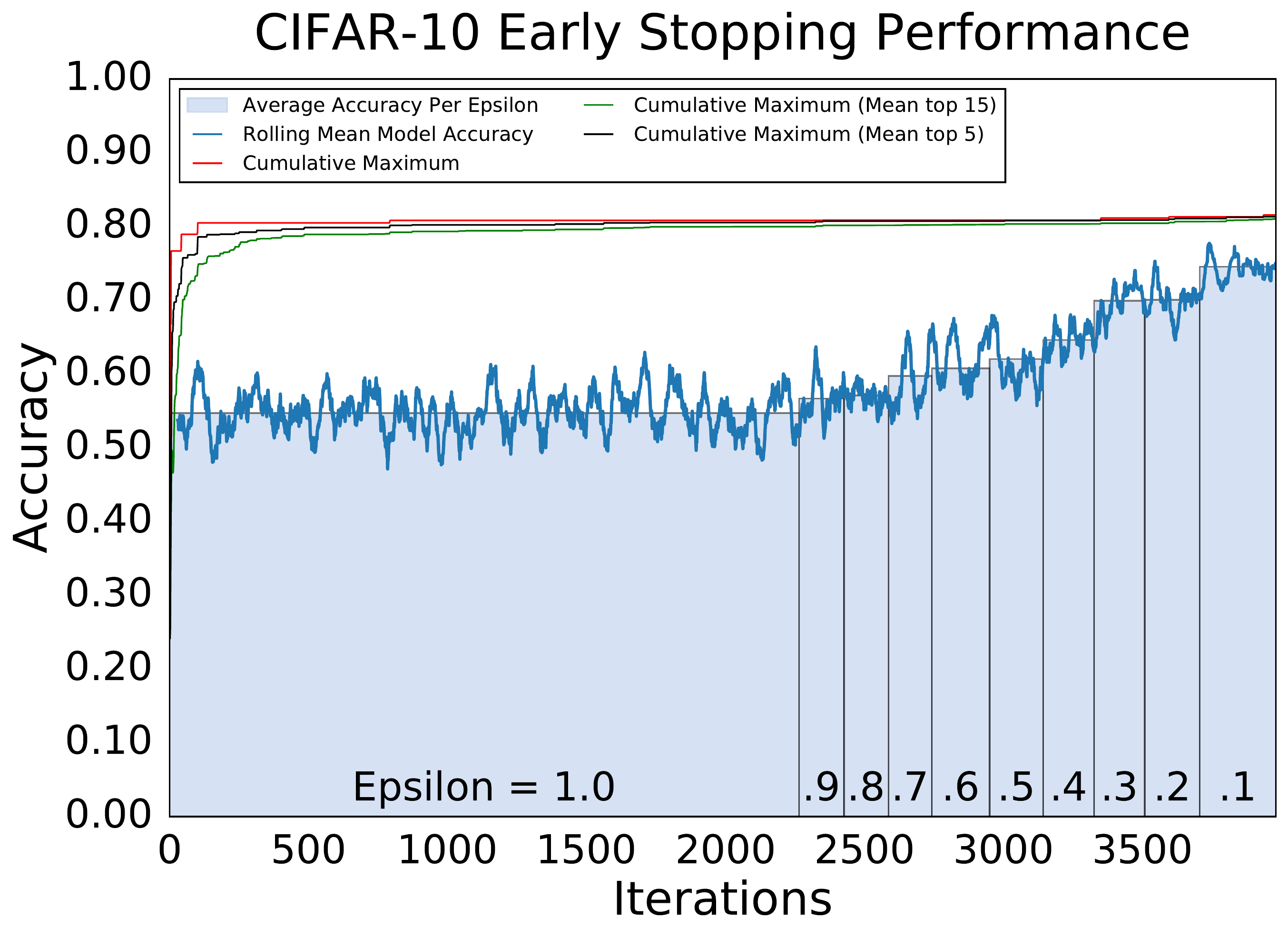}
\caption{\textbf{MetaQNN on CIFAR-10 with Early Stopping:} A full run of the MetaQNN algorithm~\citep{baker2016designing} on the CIFAR-10 dataset with early stopping. We use the $\nu$-SVR SRM with a probability threshold $\Delta = 0.99$. Light blue bars indicate the average model accuracy per decrease in $\epsilon$, which represents the shift to a more greedy policy. We also plot the cumulative best, top 5, and top 15 to show that the agent continues to find better architectures.}
\label{cifar10_metaqnn}
\end{wrapfigure}

We furthermore simulate early stopping for ResNets trained on CIFAR-10. We found that only the probability threshold $\Delta = 0.99$ resulted in recovering the top model consistently. However, even with such a conservative threshold, the search was sped up by a factor of 3.4 over the baseline. While we do not have the computational resources to run the full experiment from~\cite{zoph2016neural}, our method could provide similar gains in large scale architecture searches.

It is not enough, however, to simply simulate the speedup because meta-modeling algorithms typically use the observed performance in order to update an acquisition function to inform future sampling. In the reinforcement learning setting, the performance is given to the agent as a reward, so we also empirically verify that substituting $\hat y_T$ for $y_T$ does not cause the MetaQNN agent to converge to a subpar policy. Replicating the MetaQNN experiment on CIFAR-10 (see Figure \ref{cifar10_metaqnn}), we find that integrating early stopping with the $Q$-learning procedure does not disrupt learning and resulted in a speedup of 3.8x with $\Delta = 0.99$. The speedup is relatively low due to a conservative value of $\Delta$. After training the top models to 300 epochs, we also find that the resulting performance (just under 93\%) is on par with original results of~\cite{baker2016designing}.

\subsection{Early Stopping for Hyperparameter Optimization}
Recently,~\cite{li2016hyperband} introduced Hyperband, a random search technique based on multi-armed bandits that obtains state-of-the-art performance in hyperparameter optimization in a variety of settings. The Hyperband algorithm trains a population of models with different hyperparameter configurations and iteratively discards models below a certain percentile in performance among the population until the computational budget is exhausted or satisfactory results are obtained.

\subsubsection{Fast Hyperband}
\vspace{-5pt}
We present a Fast Hyperband (f-Hyperband) algorithm based on our early stopping scheme.  During each iteration of successive halving, Hyperband trains $n_i$ configurations to $r_i$ epochs. In f-Hyperband, we train an SRM to predict $y_{r_i}$ and do early stopping within each iteration of successive halving. We initialize f-Hyperband in exactly the same way as vanilla Hyperband, except once we have trained 100 models to $r_i$ iterations, we begin early stopping for all future successive halving iterations that train to $r_i$ iterations. By doing this, we exhibit no initial slowdown to  Hyperband due to a ``burn-in'' phase. We also introduce a parameter $\kappa$ which denotes the proportion of the $n_i$ models in each iteration that must be trained to the full $r_i$ iterations. This is similar to setting the criterion based on the $n^{\text{th}}$ best model in the previous section. See Appendix section \ref{appendix:f_hyperband} for an algorithmic representation of f-Hyperband.

\begin{figure}
\centering
\includegraphics[width=1\textwidth]{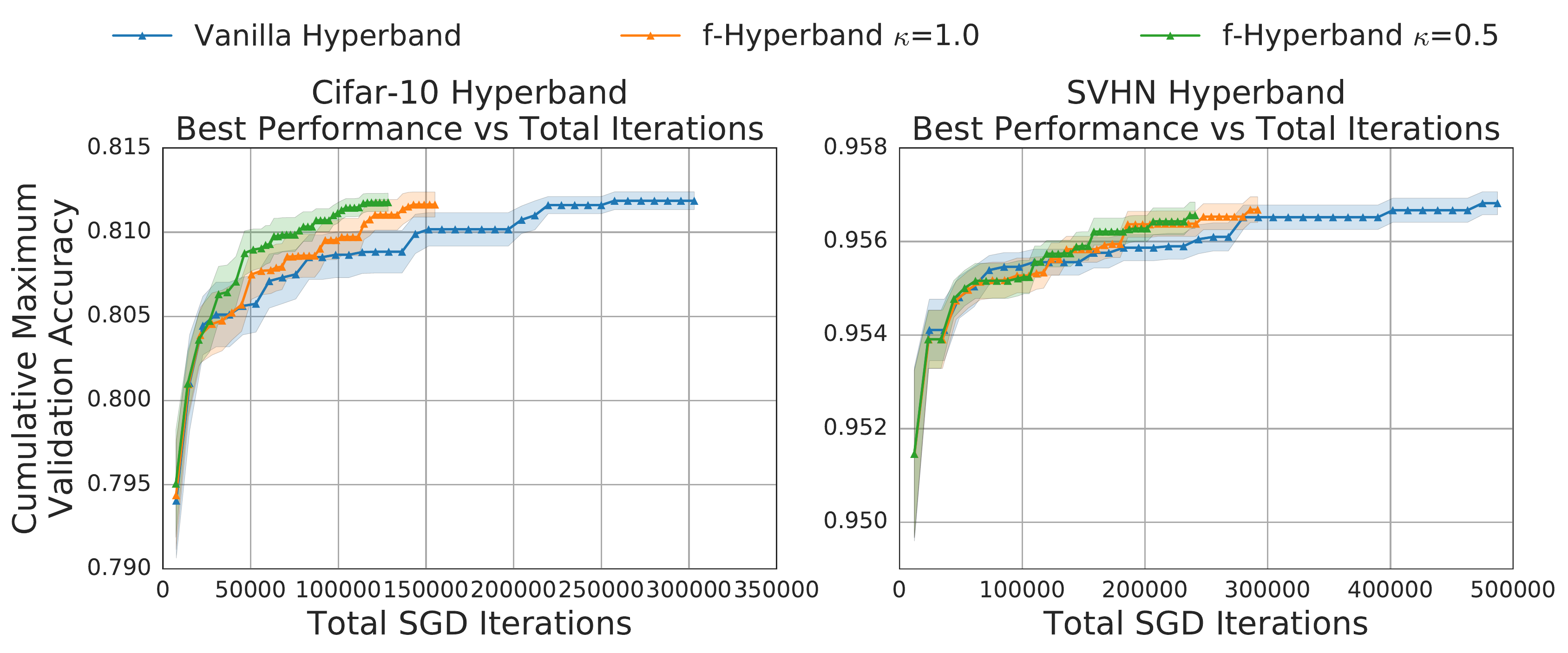}
\caption[Simulated Max Accuracy vs SGD Iterations for Hyperband]{\textbf{Simulated Max Accuracy vs SGD Iterations for Hyperband:} We show the trajectories of the maximum performance so far versus total computational resources used for 40 consecutive Hyperband runs with $\eta=3.0$ and $\Delta = 0.95$. f-Hyperband remains above the Hyperband curve at all iterations, and less aggressive settings for $\kappa$ converge to the same or better final accuracy. Each triangle marks the completion of full Hyperband algorithm.}
\label{hp_traj}
\end{figure}

We empirically evaluate f-Hyperband using Cuda-Convnet trained on CIFAR-10 and SVHN datasets. Figure \ref{hp_traj} shows that f-Hyperband evaluates the same number of unique configurations as Hyperband within half the compute time, while achieving the same final accuracy within standard error. When reinitializing hyperparameter searches, one can use previously-trained set of SRMs to achieve even larger speedups. Figure \ref{hp_speed_vs_hp} in Appendix shows that one can achieve up to a 7x speedup in such cases. 

\section{Conclusion}
\vspace{-5pt}
In this paper we introduce a simple, fast, and accurate model for predicting future neural network performance using features derived from network architectures, hyperparameters, and time-series performance data.  We show that the performance of drastically different network architectures can be jointly learned and predicted on both image classification and language models. Using our simple algorithm, we can speedup hyperparameter search techniques with complex acquisition functions, such as a $Q$-learning agent, by a factor of 3x to 6x and Hyperband---a state-of-the-art hyperparameter search method---by a factor of 2x, without disturbing the search procedure. We outperform all competing methods for performance prediction in terms of accuracy, train and test time, and speedups obtained on hyperparameter search methods. We hope that the simplicity and success of our method will allow it to be easily incorporated into current hyperparameter optimization pipelines for deep neural networks. With the advent of large scale automated architecture search~\citep{baker2016designing, zoph2016neural}, methods such as ours will be vital in exploring even larger and more complex search spaces.

\medskip
{\bibliography{neural_prediction_refs}

\begin{thebibliography}{21}
\providecommand{\natexlab}[1]{#1}
\providecommand{\url}[1]{\texttt{#1}}
\expandafter\ifx\csname urlstyle\endcsname\relax
  \providecommand{\doi}[1]{doi: #1}\else
  \providecommand{\doi}{doi: \begingroup \urlstyle{rm}\Url}\fi

\bibitem[Baker et~al.(2017)Baker, Gupta, Naik, and Raskar]{baker2016designing}
Bowen Baker, Otkrist Gupta, Nikhil Naik, and Ramesh Raskar.
\newblock Designing neural network architectures using reinforcement learning.
\newblock \emph{International Conference on Learning Representations}, 2017.

\bibitem[Bergstra \& Bengio(2012)Bergstra and Bengio]{bergstra2012random}
James Bergstra and Yoshua Bengio.
\newblock Random search for hyper-parameter optimization.
\newblock \emph{JMLR}, 13\penalty0 (Feb):\penalty0 281--305, 2012.

\bibitem[Bergstra et~al.(2013)Bergstra, Yamins, and Cox]{bergstra2013making}
James Bergstra, Daniel Yamins, and David~D Cox.
\newblock Making a science of model search: Hyperparameter optimization in
  hundreds of dimensions for vision architectures.
\newblock \emph{ICML (1)}, 28:\penalty0 115--123, 2013.

\bibitem[Brock et~al.(2017)Brock, Lim, Ritchie, and Weston]{brock2017smash}
Andrew Brock, Theodore Lim, JM~Ritchie, and Nick Weston.
\newblock Smash: One-shot model architecture search through hypernetworks.
\newblock \emph{arXiv preprint arXiv:1708.05344}, 2017.

\bibitem[Cortes et~al.(2017)Cortes, Gonzalvo, Kuznetsov, Mohri, and
  Yang]{pmlr-v70-cortes17a}
Corinna Cortes, Xavier Gonzalvo, Vitaly Kuznetsov, Mehryar Mohri, and Scott
  Yang.
\newblock {A}da{N}et: Adaptive structural learning of artificial neural
  networks.
\newblock \emph{International Conference on Machine Learning}, 70:\penalty0
  874--883, 2017.

\bibitem[Domhan et~al.(2015)Domhan, Springenberg, and
  Hutter]{domhan2015speeding}
Tobias Domhan, Jost~Tobias Springenberg, and Frank Hutter.
\newblock Speeding up automatic hyperparameter optimization of deep neural
  networks by extrapolation of learning curves.
\newblock \emph{IJCAI}, 2015.

\bibitem[Hutter et~al.(2011)Hutter, Hoos, and
  Leyton-Brown]{hutter2011sequential}
Frank Hutter, Holger~H Hoos, and Kevin Leyton-Brown.
\newblock Sequential model-based optimization for general algorithm
  configuration.
\newblock In \emph{International Conference on Learning and Intelligent
  Optimization}, pp.\  507--523. Springer, 2011.

\bibitem[Klein et~al.(2017)Klein, Falkner, Springenberg, and
  Hutter]{klein2017learning}
Aaron Klein, Stefan Falkner, Jost~Tobias Springenberg, and Frank Hutter.
\newblock Learning curve prediction with bayesian neural networks.
\newblock \emph{International Conference on Learning Representations}, 17,
  2017.

\bibitem[Krizhevsky(2012)]{cudaconv}
Alex Krizhevsky.
\newblock Cuda-convnet.
\newblock \emph{https://code.google.com/p/cuda-convnet/}, 2012.

\bibitem[Li et~al.(2017)Li, Jamieson, DeSalvo, Rostamizadeh, and
  Talwalkar]{li2016hyperband}
Lisha Li, Kevin Jamieson, Giulia DeSalvo, Afshin Rostamizadeh, and Ameet
  Talwalkar.
\newblock Hyperband: {A} novel bandit-based approach to hyperparameter
  optimization.
\newblock \emph{International Conference on Learning Representations}, 2017.

\bibitem[Negrinho \& Gordon(2017)Negrinho and
  Gordon]{negrinho2017deeparchitect}
Renato Negrinho and Geoff Gordon.
\newblock Deeparchitect: Automatically designing and training deep
  architectures.
\newblock \emph{arXiv preprint arXiv:1704.08792}, 2017.

\bibitem[Schaffer et~al.(1992)Schaffer, Whitley, and
  Eshelman]{schaffer1992combinations}
J~David Schaffer, Darrell Whitley, and Larry~J Eshelman.
\newblock Combinations of genetic algorithms and neural networks: A survey of
  the state of the art.
\newblock \emph{International Workshop on Combinations of Genetic Algorithms
  and Neural Networks}, pp.\  1--37, 1992.

\bibitem[Shahriari et~al.(2016)Shahriari, Swersky, Wang, Adams, and
  de~Freitas]{shahriari2016taking}
Bobak Shahriari, Kevin Swersky, Ziyu Wang, Ryan~P Adams, and Nando de~Freitas.
\newblock Taking the human out of the loop: A review of bayesian optimization.
\newblock \emph{Proceedings of the IEEE}, 104\penalty0 (1):\penalty0 148--175,
  2016.

\bibitem[Snoek et~al.(2012)Snoek, Larochelle, and Adams]{snoek2012practical}
Jasper Snoek, Hugo Larochelle, and Ryan~P Adams.
\newblock Practical bayesian optimization of machine learning algorithms.
\newblock \emph{NIPS}, pp.\  2951--2959, 2012.

\bibitem[Snoek et~al.(2015)Snoek, Rippel, Swersky, Kiros, Satish, Sundaram,
  Patwary, Prabhat, and Adams]{snoek2015scalable}
Jasper Snoek, Oren Rippel, Kevin Swersky, Ryan Kiros, Nadathur Satish,
  Narayanan Sundaram, Mostofa Patwary, Mr~Prabhat, and Ryan Adams.
\newblock Scalable bayesian optimization using deep neural networks.
\newblock In \emph{International Conference on Machine Learning}, pp.\
  2171--2180, 2015.

\bibitem[Stanley \& Miikkulainen(2002)Stanley and
  Miikkulainen]{stanley2002evolving}
Kenneth~O Stanley and Risto Miikkulainen.
\newblock Evolving neural networks through augmenting topologies.
\newblock \emph{Evolutionary Computation}, 10\penalty0 (2):\penalty0 99--127,
  2002.

\bibitem[Suganuma et~al.(2017)Suganuma, Shirakawa, and
  Nagao]{suganuma2017genetic}
Masanori Suganuma, Shinichi Shirakawa, and Tomoharu Nagao.
\newblock A genetic programming approach to designing convolutional neural
  network architectures.
\newblock \emph{arXiv preprint arXiv:1704.00764}, 2017.

\bibitem[Swersky et~al.(2014)Swersky, Snoek, and Adams]{swersky2014freeze}
Kevin Swersky, Jasper Snoek, and Ryan~Prescott Adams.
\newblock Freeze-thaw bayesian optimization.
\newblock \emph{arXiv preprint arXiv:1406.3896}, 2014.

\bibitem[Verbancsics \& Harguess(2013)Verbancsics and
  Harguess]{verbancsics2013generative}
Phillip Verbancsics and Josh Harguess.
\newblock Generative neuroevolution for deep learning.
\newblock \emph{arXiv preprint arXiv:1312.5355}, 2013.

\bibitem[Zoph \& Le(2017)Zoph and Le]{zoph2016neural}
Barret Zoph and Quoc~V Le.
\newblock Neural architecture search with reinforcement learning.
\newblock \emph{International Conference on Learning Representations}, 2017.

\bibitem[Zoph et~al.(2017)Zoph, Vasudevan, Shlens, and Le]{zoph2017learning}
Barret Zoph, Vijay Vasudevan, Jonathon Shlens, and Quoc~V Le.
\newblock Learning transferable architectures for scalable image recognition.
\newblock \emph{arXiv preprint arXiv:1707.07012}, 2017.

\end{thebibliography}
\bibliographystyle{iclr2018_conference}
}
\newpage
\appendix 

\section*{Appendix}

\section{Datasets and architectures}
\label{appendix:search_space}
\textbf{Deep Resnets (TinyImageNet):} We sample 500 ResNet architectures and train them on the TinyImageNet\footnote{https://tiny-imagenet.herokuapp.com/} dataset (containing 200 classes with 500 training images of $32\times32$ pixels) for 140 epochs. We vary depths, filter sizes and number of convolutional filter block outputs. Filter sizes are sampled from $\{3, 5, 7\}$ and number of filters is sampled from $\{2,3,4,...,22\}$. Each ResNet block is composed of three convolutional layers followed by batch normalization and summation layers. We vary the number of blocks from 2 to 18, giving us networks with depths varying between 14 and 110. Each network is trained for 140 epochs, using Nesterov optimizer. The learning rate is set to 0.1 and learning rate reduction and momentum are set to 0.1 and 0.9 respectively.

\textbf{Deep Resnets (CIFAR-10):} We sample 500 39-layer ResNet architectures from a search space similar to~\cite{zoph2016neural}, varying kernel width, kernel height, and number of kernels. We train these models for 50 epochs on CIFAR-10. Each architecture consists of 39 layers: 12 \textit{conv}, a 2x2 \textit{max pool}, 9 \textit{conv}, a 2x2 \textit{max pool}, 15 \textit{conv}, and \textit{softmax}. Each \textit{conv} layer is followed by batch normalization and a ReLU nonlinearity. Each block of 3 \textit{conv} layers are densely connected via residual connections and also share the same kernel width, kernel height, and number of learnable kernels. Kernel height and width are independently sampled from $\{1, 3, 5, 7\}$ and number of kernels is sampled from $\{6, 12, 24, 36\}$. Finally, we randomly sample residual connections between each block of \textit{conv} layers. Each network is trained for 50 epochs using the RMSProp optimizer, with weight decay $10^{-4}$, initial learning rate 0.001, and a learning rate reduction to $10^{-5}$ at epoch 30 on the CIFAR-10 dataset.

\textbf{MetaQNN CNNs (CIFAR-10 and SVHN):} We sample 1,000 model architectures from the search space detailed by~\cite{baker2016designing}, which allows for varying the numbers and orderings of convolution, pooling, and fully connected layers. The models are between 1 and 12 layers for the SVHN experiment and between 1 and 18 layers for the CIFAR-10 experiment. Each architecture is trained on SVHN and CIFAR-10 datasets for 20 epochs. Table \ref{state_space_table} displays the state space of the MetaQNN algorithm. 

\begin{table}[h!]
\centering
\scalebox{0.9}{
\begin{tabular}{|c|l|l|}
\hline
Layer Type & Layer Parameters & Parameter Values \\ \hline
Convolution (C) 
& \begin{tabular}[c]{@{}l@{}}$i \sim$ Layer depth \\$f \sim$ Receptive field size\\ $\ell \sim$ Stride\\ $d \sim$ \# receptive fields
\\ $n \sim$ Representation size\end{tabular} 
& \begin{tabular}[c]{@{}l@{}}$<12$\\Square. $\in\{1,3,5\}$\\   Square. Always equal to 1 \\   $\in\{64,128,256,512\}$\\ 
$\in\{(\infty, 8]$, $(8, 4]$, $(4, 1]\}$\end{tabular} \\ \hline
Pooling (P) 
& \begin{tabular}[c]{@{}l@{}}$i \sim$ Layer depth\\$(f, \ell) \sim$ (Receptive field size, Strides) \\ $n \sim$ Representation size\end{tabular} 
& \begin{tabular}[c]{@{}l@{}}$<12$\\ Square. $\in\big\{(5,3), (3,2), (2,2)  \big\}$\\ 
$\in\{(\infty, 8]$, $(8, 4]$ and $(4, 1]\}$\end{tabular} \\ \hline
Fully Connected (FC) 
& \begin{tabular}[c]{@{}l@{}} $i \sim$ Layer depth\\$n \sim$ \# consecutive FC layers\\$d \sim$ {\# neurons}\end{tabular} 
& \begin{tabular}[c]{@{}l@{}} 
$<12$\\
$<3$\\ 
$\in\{512,256,128\}$\end{tabular} \\ \hline
Termination State  
& \begin{tabular}[c]{@{}l@{}} $s \sim$  Previous State\\ $t \sim$ Type\end{tabular} 
& \begin{tabular}[c]{@{}l@{}} \\ Global Avg. Pooling/Softmax\end{tabular} \\ \hline
\end{tabular}
}

\caption{\textbf{Experimental State Space For MetaQNN.} For each layer type, we list the relevant parameters and the values each parameter is allowed to take. The networks are sampled beginning from the starting layer. Convolutional layers are allowed to transition to any other layer. Pooling layers are allowed to transition to any layer other than pooling layers. Fully connected layers are only allowed to transition to fully connected or softmax layers. A convolutional or pooling layer may only go to a fully connected layer if the current image representation size is below 8. We use this space to both randomly sample and simulate the behavior of a MetaQNN run as well as directly run the MetaQNN with early stopping.}
\label{state_space_table}
\end{table}

\textbf{LSTM (PTB):} We sample 300 LSTM models and train them on the Penn Treebank dataset for 60 epochs. Number of hidden layer inputs and lstm cells was varied from 10 to 1400 in steps of 20. Each network was trained for 60 epochs with batch size of 50 and trained the models using stochastic gradient descent. Dropout ratio of 0.5 was used to prevent overfitting. Dictionary size of 400 words was used to generate embeddings when vectorizing the data.

\textbf{Cuda-Convnet (CIFAR-10 and SVHN):} We train Cuda-Convnet architecture~\citep{cudaconv} with varying values of initial learning rate, learning rate reduction step size, weight decay for convolutional and fully connected layers, and scale and power of local response normalization layers. We train models with CIFAR-10 for 60 epochs and with SVHN for 12 epochs. Table \ref{tab:hb_params} show the hyperparameter ranges for the Cuda Convnet experiments.

\begin{table}[ht]
\begin{center}
\label{cifar_hp}
\begin{tabularx}{0.97\linewidth}{lllll}
\toprule
Experiment & Hyperparameter               & Scale   & Min  & Max \bigstrut[b]\\ \midrule
CIFAR-10, Imagenet, SVHN & Initial Learning Rate        & Log     & $5\times 10^{-5}$ & 5   \\
&Learning Rate Reductions     & Integer & 0    & 3   \\ \midrule
&Conv1 $L_2$ Penalty             & Log     & $5\times 10^{-5}$ & $5$   \\
&Conv2 $L_2$ Penalty             & Log     & $5\times 10^{-5}$ & $5$   \\
CIFAR-10, SVHN  &Conv3 $L_2$ Penalty             & Log     & $5\times 10^{-5}$ & $5$   \\
&FC4 $L_2$ Penalty               & Log     & $5\times 10^{-5}$ & $5$   \\
&Response Normalization Scale & Log     & $5\times 10^{-6}$ & $5$   \\
&Response Normalization Power & Linear  & $1\times 10^{-2}$ & $3$ \\     \bottomrule
\end{tabularx}
\vspace{0.1cm}
\caption{Range of hyperparameter settings used for the Hyperband experiment (Section 4.1)}
\label{tab:hb_params}
\end{center}
\end{table}


\section{Hyperparameter selection in Random Forest and SVM based experiments}
When training SVM and Random Forest we divided the data into training and validation and used cross validation techniques to select optimal hyperparameters. The SVM and RF model was then trained on full training data using the best hyperparameters. For random forests we varied number of trees between 10 and 800, and varied ratio of number of features from 0.1 to 0.5. For $\nu$-SVR, we perform a random search over 1000 hyperparameter configurations from the space $C \sim$ LogUniform($10^{-5}$, 10), $\nu \sim$ Uniform(0, 1), and $\gamma \sim$ LogUniform($10^{-5}$, 10) (when using the RBF kernel).
\begin{figure}[h]
\centering
\includegraphics[width=\textwidth]{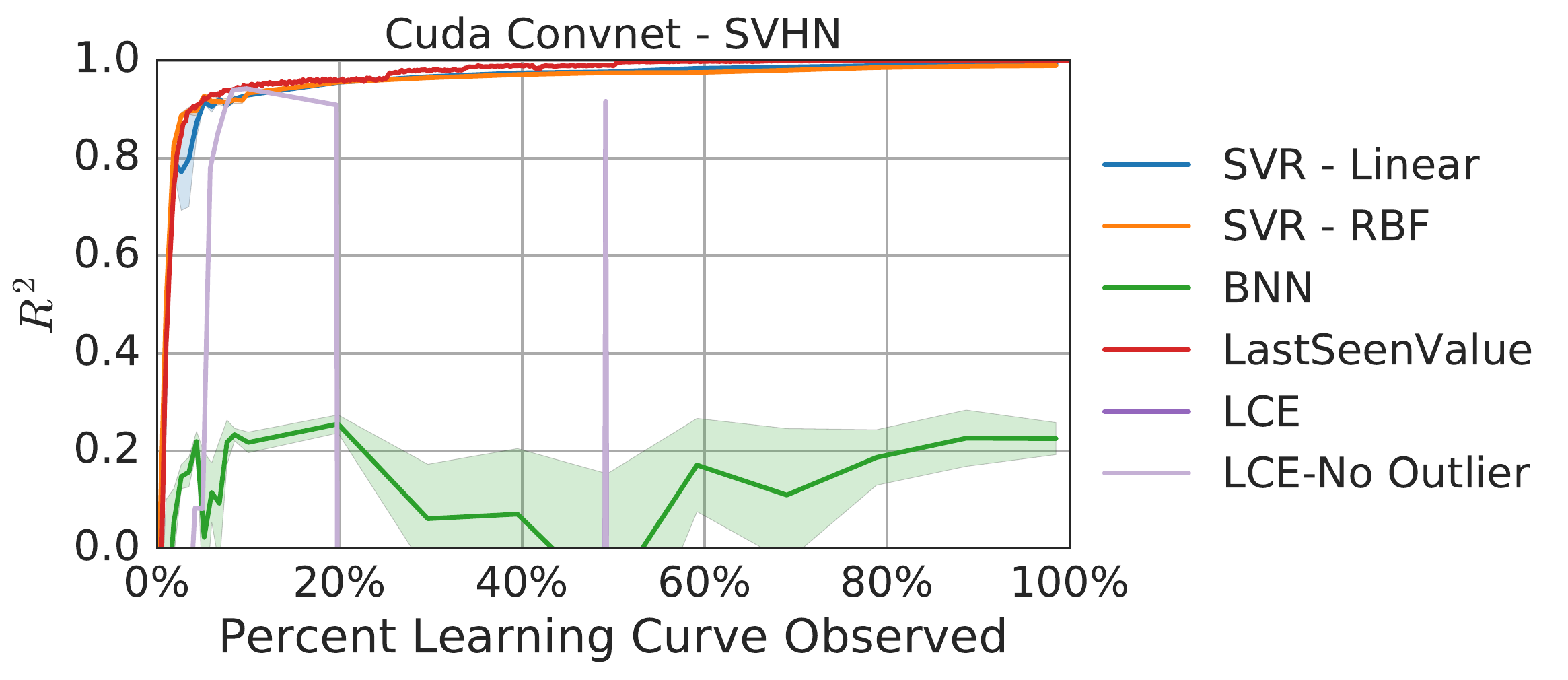}
\caption{\textbf{Cuda Convnet SVHN Performance Prediction Results:} We plot the performance of each method versus the percent of learning curve observed for the Cuda Convnet SVHN experiment. For BNN and $\nu$-SVR (linear and RBF), we sample 10 different training sets, plot the mean $R^2$, and shade the corresponding standard error. We compare our method against BNN~\citep{klein2017learning}, LCE~\citep{domhan2015speeding}, and a ``last seen value'' heuristic \citep{li2016hyperband}. Absent results for a model indicate that it did not achieve a positive $R^2$.}
\label{perf_svhn_cuda}
\end{figure}

\begin{figure}[tbh]
\centering
\includegraphics[width=0.9\textwidth]{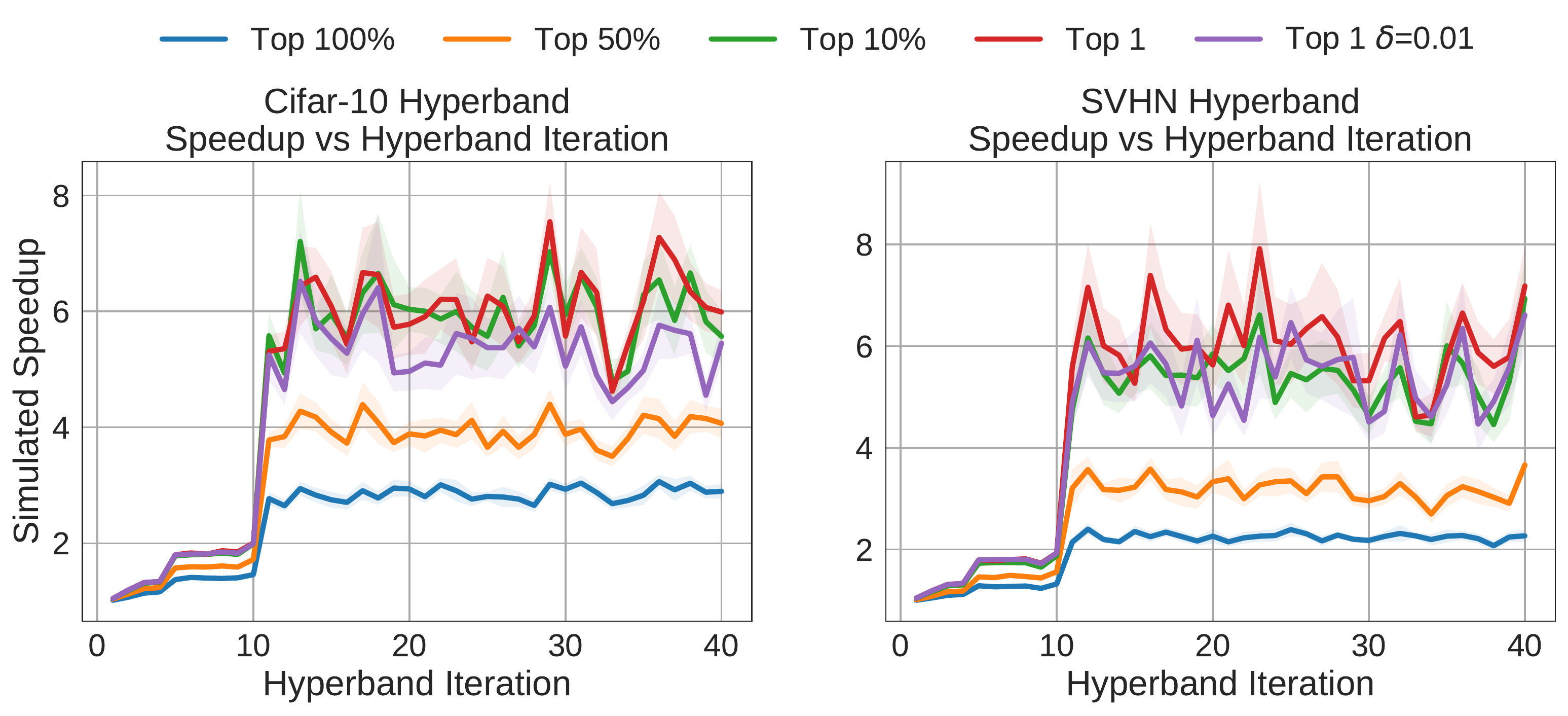}
\caption[Simulated Speedup on Hyperband vs Hyperband Iteration]{\textbf{Simulated Speedup on Hyperband vs Hyperband Iteration:} We show the speedup using the f-Hyperband algorithm over Hyperband on 40 consecutive runs with $\eta=3.0$ and $\Delta = 0.95$. The major jump in speedup comes at iteration 10, where we have trained more than 100 models to the full $R$ iterations.}
\label{hp_speed_vs_hp}
\end{figure}

\section{f-Hyperband}
\label{appendix:f_hyperband}

Algorithm \ref{alg:fhyperband} of this text replicates Algorithm 1 from \cite{li2016hyperband}, except we initialize two dictionaries: $D$ to store training data and $M$ to store performance prediction models. $D[r]$ will correspond to a dictionary containing all datasets with prediction target epoch $r$. $D[r][\tau]$ will correspond to the dataset for predicting $y_r$ based on the observed $y(t)_{1-\tau}$, and $M[r][\tau]$ will hold the corresponding performance prediction model. We will assume that the performance prediction model will have a \texttt{train} function, and a \texttt{predict} function that will return the prediction and standard deviation of the prediction. In addition to the standard Hyperband hyperparameters $R$ and $\eta$, we include $\Delta$ and $\delta$ described in Section \ref{sec:early_stop} and $\kappa$. During each iteration of successive halving, we train $n_i$ configurations to $r_i$ epochs; $\kappa$ denotes the fraction of the top $n_i$ models that should be run to the full $r_i$ iterations. This is similar to setting the criterion based on the $n^{\text{th}}$ best model in the previous section. 

We also detail the \texttt{run\_then\_return\_validation\_loss} function in Algorithm \ref{alg:run_return_val}. This algorithm runs a set of configurations, adds training data from observed learning curves, trains the performance prediction models when there is enough training data present, and then uses the models to terminate poor configurations. It assumes we have a function \texttt{max\_k}, which returns the $k^{\text{th}}$ max value or $-\infty$ if the list has less than $k$ values.

\begin{algorithm}
\caption{f-Hyperband}
\label{alg:fhyperband}
\SetKwInOut{Input}{input}
\SetKwInOut{Initialize}{initialize}
\Input{
	\begin{algotabularx}{@{}p{.03\textwidth}p{.03\textwidth}X@{}}
		$R$ &--& (Max resources allocated to any configuration) \\
		$\eta$ &--& (default $\eta =3$) \\ 
		$\Delta$ &--& (Probability threshold for early termination) \\
		$\delta$ &--& (Performance offset for early termination) \\
		$d$ &--& (\# points required to train performance predictors) \\
		$\kappa$ &--& (Proportion of models to train)
	\end{algotabularx}
}
\Initialize{
		$D = $ dict()\\ 
		$M = $ dict() \\
	    $s_\text{max} = \floor{\log_\eta (R)}$ \\
	    $B = (s_\text{max} + 1)R$
}
\For{$s \in \left\{s_\text{max}, \dots, 0\right\}$}{
     $n = \ceil{\frac{B}{R}\frac{\eta^s}{s+1}}, \hspace{10px} r= R\eta^{-s}$\\
     \texttt{// begin SUCCESSIVEHALVING with ($n$, $r$) inner loop} \\
     $T =$ \texttt{get\_hyperparameter\_configuration($n$)} \\
    \For{$i \in \left\{ 0, \dots, s \right\}$}{
         $n_i = \floor{n \eta^{-i}}$, \hspace{10px} $r_i = r \eta ^i$ \\
         $n_\text{next} = \floor{\frac{n_i}{\eta}}$ \textbf{if} $i != s$ \textbf{else} 1 \\
         $L = $ \texttt{run\_then\_return\_validation\_loss($T, r_i, n_\text{next}, D, M$)} \\
         $T = $ \texttt{top\_k($T, L, \floor{\frac{n_i}{\eta}}$)} \\
    }
    
}
\end{algorithm}

\begin{algorithm}
\caption{\texttt{run\_then\_return\_validation\_loss}}
\label{alg:run_return_val}
\SetKwInOut{Input}{input}
\SetKwInOut{Initialize}{initialize}
\Input{
	\begin{algotabularx}{@{}p{.03\textwidth}p{.03\textwidth}X@{}}
		$T$ &--& hyperparameter configurations \\
		$r$ &--& resources to use for training \\
		$n$ &--& \# configurations in next iteration of successive halving \\
		$D$ &--& dictionary storing training data \\
		$M$ &--& dictionary storing performance prediction models\\
	\end{algotabularx}
}
\Initialize{
	$L$ = \texttt{[]}
}
\For{$t \in T$}{
	$\ell$ = \texttt{[]} \\
    \For{$i \in \left\{ 0, \dots, r-1 \right\}$}{
    	 $\ell_i$ = \texttt{run\_one\_epoch\_return\_validation\_loss($t$)} \\
    	 $\ell$\texttt{.append($\ell_i$)} \\
    	\If{$M[r][i]\texttt{\upshape .trained()}$}{
    		 $\hat{y}_r$, $\sigma$ = $M[r][i]$\texttt{.predict($\ell$)} \\
    		\If{$\Phi(\texttt{\upshape max\_k($L$, $\kappa n$)} - \delta; \hat{y}_r, \sigma) \ge \Delta$}{
    			 $L$\texttt{.append($\hat{y}_r$)}\\
    			 \texttt{break}\\
    		}
    	}
    	\ElseIf{$i == r-1$}{
    		 $L$\texttt{.append($\ell_i$)}\\
    	}
    }
    \If{$\texttt{\upshape length($D[r][0]$)} < d $ $\texttt{\upshape and length($\ell$)} == r$}{
		 \{$D[r][i]$\texttt{.append($\{\ell[0, \dots, i], \ell[r]\}$)}: $i \in \{0, \dots, r - 1\}$\} \\
		\If{$\texttt{\upshape not} M[r][i]\texttt{\upshape .trained()}$}{
			 $M[r][i]$\texttt{.train($D[r][i])$} \\
		}
	}
}
\Return{$L$}
\end{algorithm}

\end{document}